\documentclass[10pt,twocolumn,letterpaper]{article}

\usepackage{cvpr}              

\usepackage[accsupp]{axessibility}
\usepackage{graphicx}
\usepackage{amsmath}
\usepackage{amsfonts,amssymb}
\usepackage{booktabs}
\usepackage{multirow}

\usepackage{algorithm}
\usepackage{algorithmic}

\usepackage{bbm}

\usepackage[pagebackref,breaklinks,colorlinks,bookmarks=false]{hyperref}

\usepackage[capitalize]{cleveref}
\crefname{section}{Sec.}{Secs.}
\Crefname{section}{Section}{Sections}
\Crefname{table}{Table}{Tables}
\crefname{table}{Tab.}{Tabs.}

\def\cvprPaperID{2510}
\def\confName{CVPR}
\def\confYear{2022}

\begin{document}

\title{Towards Semi-Supervised Deep Facial Expression Recognition with An Adaptive Confidence Margin}

\author{Hangyu Li$^1$, Nannan Wang$^1$\thanks{Corresponding author} , Xi Yang$^1$, Xiaoyu Wang$^2$, Xinbo Gao$^3$\\
{$^1$}Xidian University, {$^2$}The Chinese University of Hong Kong (Shenzhen)\\
{$^3$}Chongqing University of Posts and Telecommunications\\
{\tt\small hangyuli.xidian@gmail.com, nnwang@xidian.edu.cn, yangx@xidian.edu.cn}\\
{\tt\small fanghuaxue@gmail.com, gaoxb@cqupt.edu.cn}
}
\maketitle

\begin{abstract}
Only parts of unlabeled data are selected to train models for most semi-supervised learning methods, whose confidence scores are usually higher than the pre-defined threshold (\ie, the confidence margin). We argue that the recognition performance should be further improved by making full use of all unlabeled data. In this paper, we learn an \textbf{Ada}ptive \textbf{C}onfidence \textbf{M}argin (Ada-CM) to fully leverage all unlabeled data for semi-supervised deep facial expression recognition. All unlabeled samples are partitioned into two subsets by comparing their confidence scores with the adaptively learned confidence margin at each training epoch: (1) subset I including samples whose confidence scores are no lower than the margin; (2) subset II including samples whose confidence scores are lower than the margin. For samples in subset I, we constrain their predictions to match pseudo labels. Meanwhile, samples in subset II participate in the feature-level contrastive objective to learn effective facial expression features. We extensively evaluate Ada-CM on four challenging datasets, showing that our method achieves state-of-the-art performance, especially surpassing fully-supervised baselines in a semi-supervised manner. Ablation study further proves the effectiveness of our method. The source code is available at \url{https://github.com/hangyu94/Ada-CM}. 

\end{abstract}

\section{Introduction}
\label{sec:intro}

Facial expression recognition (FER) aims to make computers understand visual emotion. Recently, the advancement of deep FER is largely promoted by large-scale labeled datasets, \eg, RAF-DB \cite{Li_2019_TIP} and AffectNet \cite{mollahosseini2017affectnet}. However, the collection of large-scale labels is quite expensive and difficult. Besides, existing labels often fail to satisfy actual fine-grained needs and the re-labeling data requires human experts. Therefore, it is urgent to develop a powerful method for training models on a large amount of data without corresponding labels, \ie, semi-supervised deep facial expression recognition (SS-DFER).

\begin{figure}[t]
  \centering
  \setlength{\abovecaptionskip}{0.1cm}
   \includegraphics[width=1\linewidth]{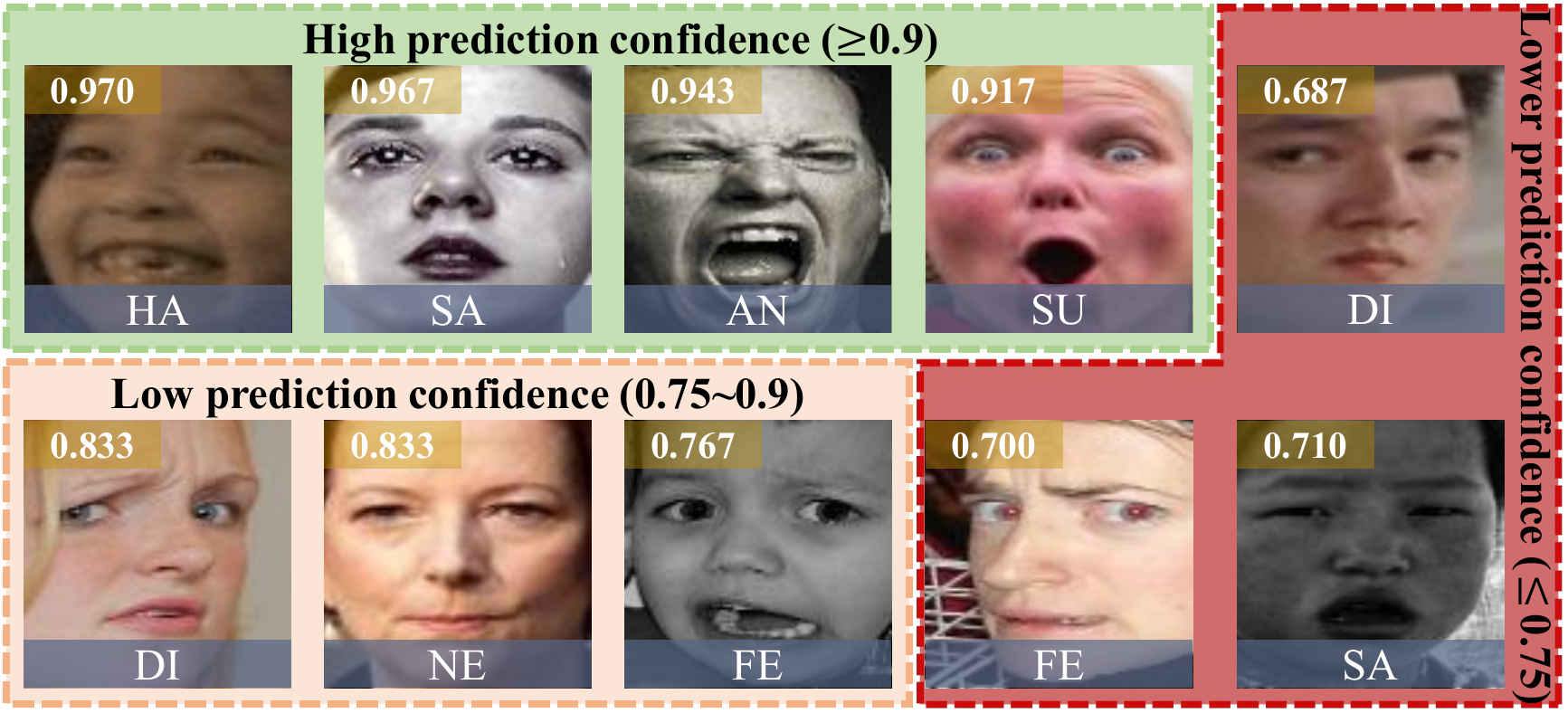}
   \caption{Confidence scores by 30 volunteers on ten faces annotated with seven classes, including \textbf{su}rprise, \textbf{fe}ar, \textbf{di}sgust, \textbf{ha}ppiness, \textbf{sa}dness, \textbf{an}ger and \textbf{ne}utral. The upper left corner of each face is tagged with its confidence score. All faces are divided into three groups based on the confidence score. The results provide insights that the confidence scores may be inconsistent among different classes and even the confidence gap between intra-class expressions may be large, \eg, faces annotated with \emph{Sadness}.}
   \vspace{-0.3cm}
   \label{fig:1}
\end{figure}

Most recent semi-supervised learning (SSL) algorithms achieve competitive performance by predicting artificial labels of unlabeled data. For example, pseudo-labeling methods \cite{lee2013pseudo,NIPS2015_378a063b,iscen2019label,xu2021dash} utilize the model predictions as artificial labels to retrain CNN models. Typically, FixMatch \cite{NEURIPS2020_06964dce} explores weakly-augmented and strongly-augmented data pairs and selects only unlabeled samples with high-confidence predictions, whose confidence scores are above the pre-defined fixed threshold (\eg, 0.95).

Despite excellent performance on common classification tasks, the threshold-based pseudo-labeling strategy is still challenging for SS-DFER mainly due to two reasons: (1) The fixed threshold for all categories. Facial expressions from different categories are classified with varying degrees of difficulty. To better understand this, we randomly pick several images from RAF-DB \cite{Li_2019_TIP} and conduct a user study. As shown in \Cref{fig:1}, for the face annotated with \emph{Happiness}, the confidence score is much higher than other facial expressions. Especially, the confidence gap between the most and least possibles is up to 28\%. Therefore, the fixed threshold is unfair to different facial expressions. In other words, the fixed threshold (\eg, 0.95) may lead to selecting too many expressions with high confidence scores (\eg, happiness) and too few expressions with low or lower confidence scores (\eg, disgust). Moreover, the fixed setting is not adaptive enough at each training epoch. (2) Inefficient data utilization. There is a large gap between the confidence scores of different intra-class samples. For example, the confidence gap between faces annotated with \emph{Sadness} is as large as 25\% (see \Cref{fig:1}). This issue may cause that some intra-class samples with low confidence scores cannot be selected for training models, \eg, the \emph{Sadness} with the confidence score of 0.71. This inspires us to consider that how samples with low confidence scores contribute to feature learning. Hence, \textbf{to fully leverage unlabeled data with the adaptive threshold is crucial for SS-DFER.}

To this end, we propose a semi-supervised DFER algorithm with an \textbf{Ada}ptive \textbf{C}onfidence \textbf{M}argin (Ada-CM) to enjoy its adaptivity in terms of the learning on all unlabeled data. Specifically, the proposed Ada-CM firstly runs over all given labeled data and adaptively updates the confidence margin based on the learning difficulty of different facial expressions. Importantly, the confidence margin is gradually improved over training epochs. Then, it predicts confidence scores of weakly-augmented unlabeled data, which are compared with the learned confidence margin to partition all unlabeled samples into two subsets: subset I including samples with high confidence scores (\ie, whose confidence scores are not lower than the margin) and subset II including samples with low confidence scores (\ie, whose confidence scores are lower than the margin). For samples in subset I, Ada-CM leverages strongly-augmented unlabeled samples and pseudo labels from their weakly-augmented versions to calculate the cross-entropy loss. Moreover, for subset II, we conduct a feature-level contrastive objective to learn effective features by applying the InfoNCE loss \cite{chen2020simple}. Overall, our main contributions can be summarized as follows:

\begin{itemize}
\item We propose a novel end-to-end semi-supervised DFER method by adaptively learning the confidence margin. To the best of our knowledge, this is the first solution to explore the dynamic confidence margin in SS-DFER.
\item An adaptive confidence margin is designed to dynamically learn on all unlabeled data for the model's training. More importantly, samples with low confidence scores are leveraged to enhance the feature-level similarity.
\item Extensive experiments on four challenging datasets show the effectiveness of our proposed Ada-CM. Especially, our method achieves superior performance, surpassing fully-supervised baselines in a semi-supervised manner.
\end{itemize}

\section{Related Work}
\label{sec:related}

\subsection{Facial Expression Recognition}

Numerous FER methods \cite{Li_2019_TIP,li2021tip,she2021dive,Xue_2021_ICCV} have been proposed. There are two major lines of research on FER, \ie, handcraft features and deep learning-based methods.

Traditionally, early attempts \cite{hu2008multi,pietikainen2011computer,luo2013facial} focus on the texture information on in-the-lab FER datasets, \eg, CK+ \cite{lucey2010extended} and Oulu-CASIA \cite{zhao2011facial}. Motivated by large-scale unconstrained FER datasets \cite{Li_2019_TIP,mollahosseini2017affectnet,BarsoumICMI2016}, DFER algorithms design effective CNN networks or loss functions to achieve superior performance. Right from the beginning, Li \textit{et al.} \cite{Li_2019_TIP} proposed a locality preserving loss to learn more discriminative facial expression features. Inspired by the attention mechanism, Wang \textit{et al.} \cite{wang2020region} proposed region-based attention network to capture important facial regions. Li \textit{et al.} \cite{li2019occlusion} explored partially-occluded facial expression recognition. Moreover, several works \cite{zeng2018facial,wang2020suppressing,she2021dive} considered the inconsistent annotation problem in DFER. Besides, Xue \textit{et al.} \cite{Xue_2021_ICCV} first explored relation-aware representations for Transformers-based DFER. 

The above methods perform FER in a fully-supervised manner. Differently, Florea \textit{et al.} \cite{margin2020eccv} proposed an extension of MixMatch \cite{NEURIPS2019_1cd138d0}, namely Margin-Mix, and leveraged unlabeled samples to solve the dense area problem. Indeed, Margin-Mix determined artificial labels of unlabeled samples by the embeddings for class centers, not by the confidence margin. Moreover, the center updating is costly and time-consuming. To the best of our knowledge, no threshold-based pseudo-labeling method has been proposed for the SS-DFER task. In our work, an adaptive confidence margin is designed to produce high-quality pseudo labels of unlabeled samples with high confidence scores.

\subsection{Semi-Supervised Learning}

In recent years, semi-supervised learning methods have been successfully applied to some challenging problems \cite{NEURIPS2020_06964dce,zhang2021flexmatch,wang2021data}. Existing works on SSL deploy consistency regularization \cite{sajjadi2016regularization,NEURIPS2020_44feb009}, entropy minimization \cite{grandvalet2004semi,lee2013pseudo} and traditional regularization \cite{NEURIPS2019_1cd138d0} to leverage unlabeled data. 

\begin{figure*}[t]
  \centering
  \setlength{\abovecaptionskip}{0.1cm}
   \includegraphics[width=1\linewidth]{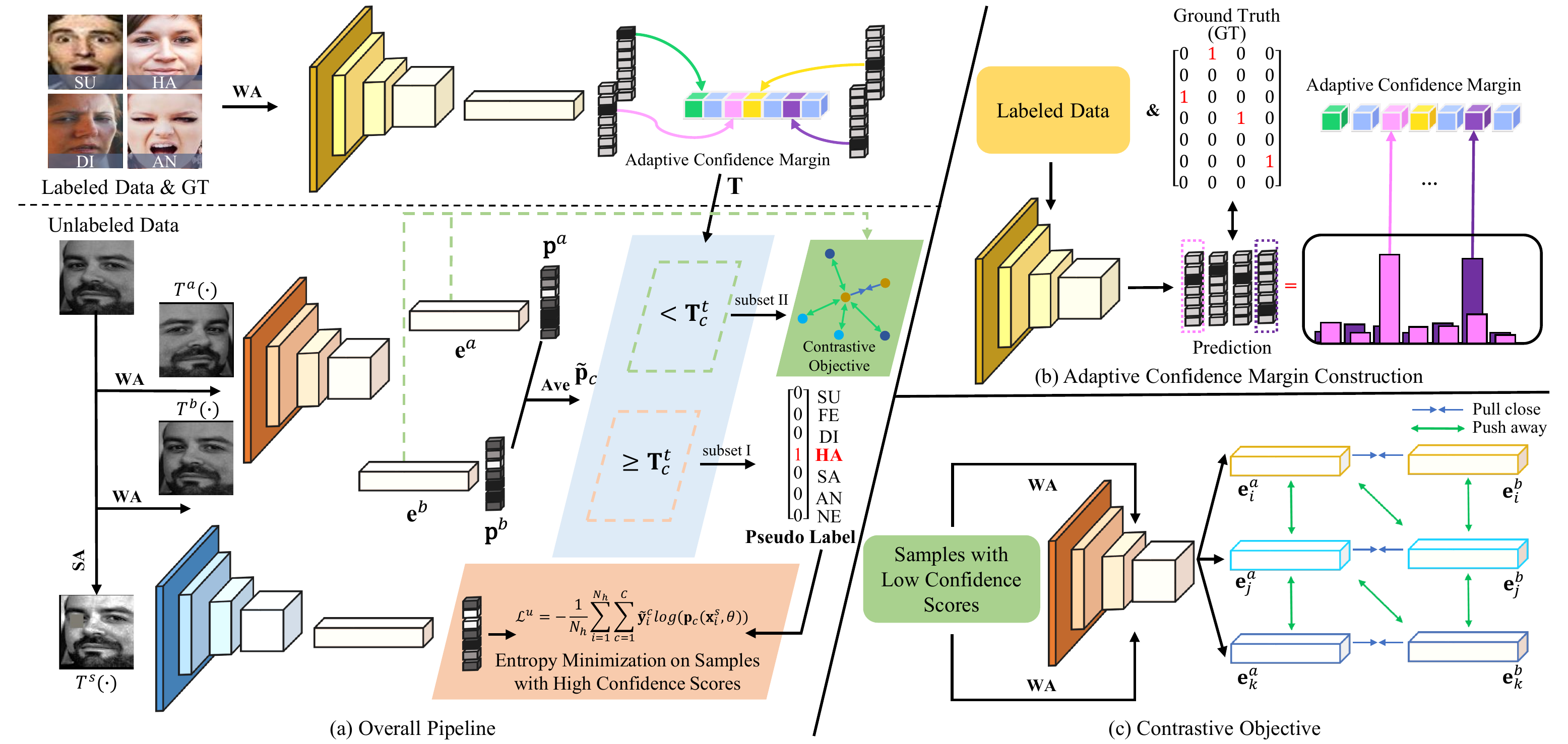}
   \caption{Illustration of Ada-CM. In each forward pass, weakly-augmented (WA) labeled samples are fed into the model to learn the adaptive confidence margin. Specifically, when the model's prediction is equal to the ground truth, the corresponding confidence scores are put into the confidence margin and then the average is used as the learned margin. Next, two WA unlabeled samples are fed separately into the model, resulting in probability distributions $\mathbf{p}^{a}$ and $\mathbf{p}^{b}$. Then, Ada-CM partitions all unlabeled data into two subsets based on the relationship between the confidence score (\ie, the maximum value in the average probability distribution $\mathbf{\widetilde{p}}_{c}$) and the confidence margin $\mathbf{T}_{c}^{t}$. Finally, samples in subset I with pseudo labels and the feature similarity on samples in subset II are explored by entropy minimization and contrastive objective, respectively. For clarity, we present the same model with three colors to distinguish different inputs.}
   \label{fig:2}
\end{figure*}

Among them, pseudo-labeling is a pioneer SSL method to obtain hard labels from model predictions. Especially, the threshold-based methods \cite{ssl_conf_05,NEURIPS2020_06964dce} select unlabeled samples with high-confidence predictions. FixMatch \cite{NEURIPS2020_06964dce} and UDA \cite{NEURIPS2020_44feb009} obtained pseudo labels based on the fixed threshold and leveraged weak and strong augmentations to achieve the consistency regularization. In addition, several works have investigated on the dynamic threshold \cite{xu2021dash,zhang2021flexmatch}. For example, Xu \textit{et al.} \cite{xu2021dash} proposed a generic method to dynamically select samples with high-confidence predictions. In our work, it is the first attempt to learn an adaptive confidence margin for SS-DFER. Besides, all unlabeled samples are learned, which is also the first attempt for SSL.

\section{Method}
\label{sec:method}

\subsection{Problem Formulation}

Generally, for a $C$-class fully-supervised DFER task, there is a set of instance-label pairs as $\mathcal{C}=(\mathcal{X},\mathcal{Y})$, where $\mathcal{X}=\{\mathbf{x}_{i}\}_{i=1}^{N}$ and $\mathcal{Y}=\{\mathbf{y}_{i}\in\{0,1\}^{C}\}_{i=1}^{N}$ are the set of training data and the corresponding one-hot labels, and $N$ denotes the number of labeled training data. The conventional loss function is the cross-entropy loss on the labeled training data:

\begin{equation}
  \mathcal{L}_{CE}^{s} = -\frac{1}{N}\sum_{i=1}^{N}\sum_{c=1}^{C}\mathbf{y}_{i}^{c}log(\mathbf{p}_{c}(\mathbf{x}_{i},\theta)),
  \label{eq:1}
\end{equation}
where $\mathbf{p}_{c}(\mathbf{x}_{i},\theta)$ denotes the prediction probability of data $\mathbf{x}_{i}$ for class $c$ with the model parameter $\theta$. However, for the problem of semi-supervised DFER, labels are not guaranteed to be fully available. In general, the original training samples are partitioned into two sets, including a labeled set and an unlabeled set. Let 
\begin{equation}
  \mathcal{S} = \{(\mathbf{x}_{i},\mathbf{y}_{i}),i=1,...,N_{s}\}
  \label{eq:2}
\end{equation}
be the labeled training set. $N_{s}$ is the number of labeled training data. Besides the labeled set $\mathcal{S}$, the unlabeled training set shares the same categories, denoted by 
\begin{equation}
  \mathcal{U} = \{\mathbf{x}^{u}_{i},i=1,...,N_{u}\},
  \label{eq:3}
\end{equation}
where $N_{u}$ is the number of unlabeled training data. 

Given the above data, existing pseudo-labeling methods \cite{lee2013pseudo,NEURIPS2020_44feb009,NEURIPS2020_06964dce} aim to generate the pseudo label $\mathbf{\widetilde{y}}_{i}$ for a sample $\mathbf{x}_{i}^{u}$. Then, the model is optimized on the labeled set $\mathcal{S}$ and the unlabeled set $\mathcal{U}$ with pseudo labels by the cross-entropy loss. For example, FixMatch \cite{NEURIPS2020_06964dce} adopts a fixed threshold for all categories and selects unlabeled data with high-confidence predictions whose confidence scores are above the threshold. Crucially, for the consistency regularization \cite{NEURIPS2019_1cd138d0} in SSL, FixMatch conducts two separate weakly-augmented (WA) and strongly-augmented (SA) operations and estimates pseudo labels based on the WA data.\footnote{It is a form of consistency regularization in which the model should output the same prediction for the WA and SA data.}

More importantly, the quality of pseudo labels depends on the threshold, which can determine the level of confidence scores. However, existing methods can only make sure that samples with high confidence scores are used for the model's training. In addition, many facial expressions (\eg, happiness) usually have higher confidence scores than certain facial expressions, which is unfair to other categories. In this work, we focus on the confidence margin-based pipeline and leverage all unlabeled data regardless of the degree of confidence scores.

\subsection{Our proposed Ada-CM} 

In this section, we first present the overall framework in \cref{sec:overall}. In \cref{sec:threshold}, we propose an adaptive confidence margin, which contains different thresholds for facial expression categories. Furthermore, we introduce the learning on all unlabeled data in \cref{sec:selection}. Finally, we display the whole training objective in \cref{sec:function}.

\subsubsection{The Overall Framework}
\label{sec:overall}

To fully leverage unlabeled data, we propose a semi-supervised DFER method (see \Cref{fig:2}). Unlike the fixed threshold for all categories, we propose an adaptive confidence margin (Ada-CM), which consists of different thresholds for each facial expression category. Then, our Ada-CM partitions all unlabeled data into two subsets by comparing the confidence scores\footnote{For labeled and unlabeled data, the confidence score can be viewed as the probability value corresponding to the ground truth and the maximum value in a probability distribution, respectively.} with the margin. Once the confidence score of unlabeled data (\ie, the maximum value in the average probability distribution $\mathbf{\widetilde{p}}_{c}$) is no lower than the corresponding threshold in the margin, the prediction on the SA version will match the pseudo label from the above WA versions via the cross-entropy loss. Otherwise, the contrastive objective is used to enhance the similarity between two WA features. Therefore, our Ada-CM mainly contains two components, including learning an adaptive confidence margin and adaptively learning on unlabeled data. We will elaborate on key technologies in turn.

\subsubsection{Adaptive Confidence Margin} 
\label{sec:threshold}

Recent SSL progresses \cite{lee2013pseudo,NEURIPS2020_44feb009,NEURIPS2020_06964dce} select unlabeled samples with high confidence scores to update models based on a  fixed threshold for all categories. However, since the confidence score varies by category, it is unfair to different facial expressions. Motivated by this, we aim to evaluate the confidence margin based on given labeled data and build an adaptive confidence margin. Note that our method requires no extra labeled data to determine the margin. 

For the labeled set $\mathcal{S} = \{(\mathbf{x}_{i},\mathbf{y}_{i}),i=1,...,N_{s}\}$, we would like to explore the confidence margin for different facial expressions. A classical idea is to obtain the predictions of all labeled samples and calculate different thresholds by averaging the corresponding confidence scores. However, this practice shows a fatal problem for SS-DFER. In particular, several studies have shown that noisy labels exist in DFER datasets \cite{wang2020suppressing,she2021dive}, which indicates that certain confidence scores of samples are not desirable. Therefore, we propose the adaptive confidence margin based on the \emph{correct} confidence scores. 


Specifically, we first obtain the predictions of all labeled samples and determine predicted labels. Compared to the ground truth $\{\mathbf{y}_{i}\in\{0,1\}^{C},i=1,2,...,N_{s}\}$, we pick out the correctly-predicted samples $\mathcal{S}^{T} = \{(\mathbf{x}_{i},\widehat{\mathbf{y}}_{i},s_{i}),i=1,...,N_{st}\}$, where $s_{i}$ is the confidence score of the $i$-th labeled data, $\widehat{\mathbf{y}}_{i}\in\{1,2,...,C\}$ denotes the $i$-th label and $N_{st}$ is the number of data in $\mathcal{S}^{T}$. We then construct the adaptive confidence margin $\mathbf{T}=\{(\mathbf{T}_{1},...,\mathbf{T}_{C})|\mathbf{T}_{c}\in\mathbb
{R},c=1,...,C\}$ by

\begin{equation}
  \mathbf{T}_{c} = \frac{1}{N_{st}^{c}}\sum_{i=1}^{N_{st}}\mathbbm{1}(\widehat{\mathbf{y}}_{i}=c)\cdot s_{i},
  \label{eq:4}
\end{equation} 
where $N_{st}^{c}$ reflects the number of samples annotated with the $c$-th class in $\mathcal{S}^{T}$. It is known that with the increasing of epoch, the discriminative ability of DFER model is stronger. Therefore, we consider that the confidence margin is also adaptively improved with the training epoch. Mathematically, the confidence margin at epoch $t$ is given by 

\begin{equation}
  \mathbf{T}_{c}^{t} =  \frac{B\mathbf{T}_{c}}{1+\gamma^{-t}},
  \label{eq:5}
\end{equation} 
where $0<B<1$ and $\gamma>1$ are two constants. In practice, we set $B = 0.97$ to control too large margin. Moreover, we use $\gamma = e$ as the default setting. The ablation study about $B$ and $\gamma$ will be shown in \cref{sec:ablation}.

\subsubsection{Adaptively Learning on Unlabeled Data}
\label{sec:selection}

The proposed adaptive confidence margin is an important criterion to determine the level of confidence scores. To leverage all unlabeled samples efficiently, we design an adaptive learning strategy to explore all unlabeled data for updating model parameters. 

To this end, we propose to adaptively learn on all unlabeled data based on the above adaptive confidence margin. Specifically, we first generate two WA versions $\mathbf{x}^{a}_{i}=T^{a}(\mathbf{x}_{i}^{u})$ and  $\mathbf{x}^{b}_{i}=T^{b}(\mathbf{x}_{i}^{u})$ and utilize the same model to extract facial expression features and probability distributions. Based on two probability distributions $\mathbf{p}^{a}$ and $\mathbf{p}^{b}$, we then compute the average probability distribution:

\begin{equation}
  \mathbf{\widetilde{p}}_{c} = \frac{1}{2}(\mathbf{p}^{a}(\mathbf{x}_{i}^{a},\theta)+\mathbf{p}^{b}(\mathbf{x}_{i}^{b},\theta)),
  \label{eq:6}
\end{equation}
where $\mathbf{\widetilde{p}}_{c}$ denotes the probability distribution of data $\mathbf{x}_{i}^{u}$ about class $c$. Now, the adaptive learning strategy compares two values, \ie, $max(\mathbf{\widetilde{p}}_{c})$ and $\mathbf{T}^{t}_{\mathop{argmax}\limits_{c}\mathbf{\widetilde{p}}_{c}}$, to dynamically partition all unlabeled data into the subset I including samples with high confidence scores and the subset II including samples with low confidence scores. 

For samples in subset I, we retain the average as the pseudo label at the current epoch, \ie, $\mathbf{\widetilde{y}}_{i}=\mathop{argmax}\limits_{c}\mathbf{\widetilde{p}}_{c}$, where $\mathbf{\widetilde{y}}_{i}$ denotes the one-hot label for convenience. To achieve the consistency regularization, we adopt the strongly-augmented operations and make the prediction of SA version match the pseudo label obtained from two WA versions. Therefore, given a high-confidence sample $\mathbf{x}_{i}^{u}$, the unsupervised loss $\mathcal{L}^{u}$ is defined as the cross-entropy loss between the SA version $\mathbf{x}_{i}^{s}=T^{s}(\mathbf{x}_{i}^{u})$ and $\mathbf{\widetilde{y}}_{i}$:

\begin{equation}
  \mathcal{L}^{u} = -\frac{1}{N_{h}}\sum_{i=1}^{N_{h}}\sum_{c=1}^{C}\mathbf{\widetilde{y}}_{i}^{c}log(\mathbf{p}_{c}(\mathbf{x}_{i}^{s},\theta)),
  \label{eq:7}
\end{equation}
where $N_{h}$ denotes the number of data in subset I. 

For samples in subset II, since the low-confidence predictions are not convincing, the cross-entropy loss cannot be used to guide the model's learning. Inspired by contrastive learning \cite{chen2020simple,Li_Peng_2021,yao2021jo}, we consider the relationship between two WA versions of the same unlabeled data to improve the discriminative power of facial expression features. Specifically, the feature-level similarity is first measured by 

\begin{equation}
    s(\mathbf{e}^{a}_{i},\mathbf{e}^{b}_{i})=\frac{(\mathbf{e}^{a}_{i})(\mathbf{e}^{b}_{i})^\top}{||\mathbf{e}^{a}_{i}||||\mathbf{e}^{b}_{i}||},
\label{eq:8}
\end{equation}
where $\mathbf{e}^{a}_{i}$ and $\mathbf{e}^{b}_{i}$ are two weak-augmented facial expression features. Based on the obtained similarity measure, the contrastive objective for the feature $\mathbf{e}^{a}_{i}$ of a sample $\mathbf{x}_{i}^{u}$ can be defined as follows:

\begin{equation}
    \mathcal{L}^{c}=-\frac{1}{N_{l}}\sum_{i=1}^{N_{l}}\log\left(\frac{e^{s(\mathbf{e}^{a}_{i},\mathbf{e}^{b}_{i}) / \tau}}{\sum\limits_{j} e^{s(\mathbf{e}^{a}_{i},\mathbf{e}^{a}_{j}) / \tau} +\sum\limits_{k} e^{s(\mathbf{e}^{a}_{i},\mathbf{e}^{b}_{k}) / \tau}}\right),
\label{eq:9}
\end{equation}
where $i,k\in I=\{1,2,3,...,N_{l}\}$, $j\in I \backslash \{i\}$, $N_{l}$ is the number of data in subset II and $\tau$ is a temperature parameter to control the softness \cite{hinton2015distilling}. Notably, this process can further increase the discriminative power of features and introduce no additional trainable parameters.

\subsubsection{Overall Objective Function} 
\label{sec:function}

The proposed SS-DFER method with the adaptive confidence margin is optimized in the end-to-end process. The whole network minimizes the following loss function:

\begin{equation}
  \mathcal{L}_{total} = \lambda_{1}\mathcal{L}^{s}_{CE} + \lambda_{2}\mathcal{L}^{u} + \lambda_{3}\mathcal{L}^{c},
  \label{eq:10}
\end{equation}
where $\mathcal{L}^{s}_{CE}$ and $\mathcal{L}^{u}$ denote the cross-entropy loss on labeled samples and unlabeled samples in subset I, respectively. $\mathcal{L}^{c}$ denotes the contrastive objective on samples in subset II. $\lambda_{1}$, $\lambda_{2}$ and $\lambda_{3}$ are hyper-parameters to balance each term's intensity. The whole process of our proposed method is summarized in \cref{alg1}.

\begin{algorithm}[t]
   \caption{Ada-CM's main learning algorithm.}
   \label{alg1}
\begin{algorithmic}[1]
   \begin{small}
   \REQUIRE Model parameters $\theta$, labeled samples and their labels $\mathcal{S} = \{(\mathbf{x}_{i},\mathbf{y}_{i}),i=1,...,N_{s}\}$, unlabeled samples $\mathcal{U}=\{\mathbf{x}^{u}_{i},i=1,...,N_{u}\}$, number of epoch $t_{max}$ and learning rate $\eta$.
   \ENSURE Updated model parameters $\theta$.
   \STATE  \textcolor[RGB]{175,171,171}{// Learning the adaptive confidence margin.}
   \STATE Initialization: $\mathbf{T}^{0}\in\{f\}^{C}$.
   \FOR{$i=1,2,3,...,N_{s}$}
   \STATE Obtain the correctly-predicted set $\mathcal{S}^{T}$.
   \STATE Update $\mathbf{T}_{c}$ by \cref{eq:4}.
   \ENDFOR
   \STATE Obtain the current confidence margin $\mathbf{T}_{c}^{t}$ by \cref{eq:5}.
   \STATE  \textcolor[RGB]{175,171,171}{// Training models with labeled and unlabeled samples.}
   \FOR{$t=1,2,3,...,t_{max}$}
   \STATE Compute $\mathcal{L}^{s}_{CE}$ using labeled samples by \cref{eq:1}.
   \STATE Predict $\mathbf{p}^{a}$, $\mathbf{p}^{b}$ and the average $\mathbf{\widetilde{p}}_{c}$ by \cref{eq:6}.
   \IF{$max(\mathbf{\widetilde{p}}_{c})\geq\mathbf{T}^{t}_{\mathop{argmax}\limits_{c}\mathbf{\widetilde{p}}_{c}}$}
   \STATE Compute $\mathcal{L}^{u}$ using subset I by \cref{eq:7}.
   \STATE Update $\theta \leftarrow \theta-\eta \nabla \mathcal{L}^{u}$.
   \ELSE
   \STATE Compute $\mathcal{L}^{c}$ using subset II by \cref{eq:8,eq:9}.
   \STATE Update $\theta \leftarrow \theta-\eta \nabla \mathcal{L}^{c}$.
   \ENDIF
   \STATE Update $\theta \leftarrow \theta-\eta \nabla \mathcal{L}^{s}_{CE}$.
   \ENDFOR
   \end{small}
\end{algorithmic}
\end{algorithm}

\subsection{Discussion}
\label{sec:discuss}

Here, we discuss the relations between the proposed Ada-CM, FixMatch \cite{NEURIPS2020_06964dce}, Dash \cite{xu2021dash} and FlexMatch \cite{zhang2021flexmatch}, which share similar philosophy but with different roles. 




\emph{Relation to FixMatch} \cite{NEURIPS2020_06964dce}. FixMatch focuses on the fixed threshold so its modeling capacity is limited at the early training stage \cite{xu2021dash,zhang2021flexmatch}. Ada-CM aims at the adaptive confidence margin, which is friendly for the early training. In addition, FixMatch only selects unlabeled samples with high confidence scores through the fixed threshold for all categories, while Ada-CM leverages all unlabeled data and learns dynamic thresholds for different facial expressions.

\emph{Relation to Dash} \cite{xu2021dash}. Dash devotes to selecting unlabeled samples whose loss values are smaller than the dynamic threshold. However, Ada-CM leverages all unlabeled samples and compares the confidence score, which intuitively reflects the predictions of unlabeled samples. Furthermore, Ada-CM is built on correctly-predicted labeled data for different categories, while Dash leverages the entire labeled set to obtain a dynamic threshold for all categories. 

\emph{Relation to FlexMatch} \cite{zhang2021flexmatch}. FlexMatch first considers the learning difficulties of each category but only selects unlabeled data with high confidence scores. In addition, FlexMatch obtains dynamic thresholds for different categories based on the number of \emph{unlabeled data} whose predictions fall into this category and above the fixed threshold. While our Ada-CM is decided by the average confidence scores of correctly-predicted \emph{labeled data} in different categories.

\section{Experiments}
\label{sec:experiments}

In this section, extensive experiments are conducted to verify the effectiveness of our proposed method. We first briefly introduce the experiment setup (\cref{sec:setup}). Then, we perform the ablation study (\cref{sec:ablation}) to show the importance of each component in Ada-CM. Finally, we compare our method with state-of-the-art methods (\cref{sec:sota,sec:cross,sec:visual}). 

\subsection{Experiment Setup}
\label{sec:setup}

{\bf Datasets.} We evaluate Ada-CM on four commonly used datasets: RAF-DB, SFEW, AffectNet and CK+. {\bf RAF-DB} \cite{Li_2019_TIP} includes nearly 30,000 facial images with two different subsets by 40 annotators. In our experiments, we choose the single-label subset with six basic expressions (\ie, surprise, fear, disgust, happy, sad and anger) and the neutral face, which is divided into the training set and testing set with the size of 12,271 and 3,068, respectively. {\bf SFEW} \cite{dhall2011static} is a static facial expression dataset selected from movies, including 958 images for training, 436 images for validation and 372 images for testing. The images in SFEW are annotated with six basic expressions and the neutral face as in RAF-DB. For the reason that there are no public labels in the testing set, we compare performance on the validation set. {\bf AffectNet} \cite{mollahosseini2017affectnet} is currently the largest real-world facial expression dataset, consisting of about 420,000 manually-annotated images with eight expression labels. For a fair comparison, we utilize 280,000 training images and 4,000 validation images (500 images per class). {\bf The Extended Cohn-Kanade (CK+)} \cite{lucey2010extended} includes 593 video sequences from 123 subjects. We select the first and last frame of each sequence as the neutral face and targeted expression, including 636 images with seven expression labels.

{\bf Performance Metrics.} For evaluating the model performance, we utilize the overall test accuracy as the performance metric for all algorithms. Besides, we follow the standard SSL evaluation protocol and perform experiments five times using different random seeds to obtain the mean accuracy and their standard deviations.

{\bf Implementation Details.} In the following experiments, we use MTCNN \cite{zhang2016joint} to detect and resize facial expressions with the size 224 $\times$ 224. Our proposed method is implemented with the PyTorch toolbox on two NVIDIA Tesla V100 GPUs. For the backbone CNN, we use the ResNet-18 \cite{he2016deep} pre-trained on MS-Celeb-1M face recognition dataset by default. We also conduct experiments with WideResNet-28-2 \cite{Zagoruyko2016WRN} used in MarginMix \cite{margin2020eccv} for a fair comparison. We employ a DFER-related weak augmentation strategy, including \emph{RandomCrop} and \emph{RandomHorizontalFlip}. Moreover, the RandAugment \cite{cubuk2020randaugment} is used as the strong augmentation scheme following by \cite{NEURIPS2020_06964dce}. The training data in RAF-DB is added in SFEW as additional unlabeled data. 

For a fair comparison, we use the Adam optimizer \cite{kingma2014adam} with the initial learning rate of 5 $\times$ $10^{-4}$ for all experiments. The total number of training epochs is set to 20. The mini-batch size of labeled and unlabeled data is 16 except for AffectNet. These setups are the same for all algorithms for fair comparisons. The initial threshold set is empirically set to $\mathbf{T}^{0}=\{0.8\}^{C}$. In the \cref{eq:10}, the hyper-parameters $\lambda_{1}$, $\lambda_{2}$ and $\lambda_{3}$ are set as 0.5, 1 and 0.1, respectively.

\begin{table}\footnotesize
  \centering
  \caption{Ablation study of the fixed threshold and different components in Ada-CM on RAF-DB and SFEW (in \%, mean $\pm$ standard deviation). Baseline denotes that the model is only trained by $\mathcal{L}^{s}_{CE}$ with limited labeled data. This also applies to the following tables. Note that $\mathcal{L}^{u}$ denotes different thresholds for obtaining data with high confidence scores, \eg, fixed (rows 2 to 4), dynamic (row 5) and our adaptive confidence margin (rows 6 and 8).}
  \vspace{-3.6mm}
  \begin{tabular}{c|cc|c|c}
    \toprule[1pt]
    \multirow{2}{*}{Method} &\multirow{2}{*}{$\mathcal{L}^{u}$}&\multirow{2}{*}{$\mathcal{L}^{c}$}& \multicolumn{1}{c|}{RAF-DB}& \multicolumn{1}{c}{SFEW} \\\cline{4-4} \cline{5-5}
    ~& & &100 labels &400 labels\\
    \hline
    Baseline&-&-&52.43{\tiny$\pm$2.24}&43.85{\tiny$\pm$2.83}\\
    \hline
    FT = 0.5&\checkmark&-&  57.49{\tiny$\pm$1.77} & 47.85{\tiny$\pm$1.89}\\
    FT = 0.8&\checkmark&-&  58.94{\tiny$\pm$2.05} & 48.58{\tiny$\pm$1.32}\\
    FT = 0.95&\checkmark&-& 60.67{\tiny$\pm$2.25} & 50.37{\tiny$\pm$0.45}\\
    \hline
    FlexMatch \cite{zhang2021flexmatch}&\checkmark&-& 61.23{\tiny$\pm$2.27} & 50.99{\tiny$\pm$1.45}\\
    \hline
    \multirow{3}{*}{Ada-CM}&\checkmark&-& 61.50{\tiny$\pm$2.10} & 51.04{\tiny$\pm$0.58} \\
    &-&\checkmark& 54.27{\tiny$\pm$2.79} & 45.99{\tiny$\pm$0.35}\\
    &\checkmark&\checkmark& 62.36{\tiny$\pm$1.10} & 52.43{\tiny$\pm$0.67}\\
    \bottomrule[1pt]
    \end{tabular}
    \vspace{-0.3cm}
  \label{tab:ab1}
\end{table}

\subsection{Ablation Study}
\label{sec:ablation}

In this section, we analyze the contribution of each component in our method. For convenience, we use `FT' to refer to FixMatch \cite{NEURIPS2020_06964dce} with different fixed thresholds in the following experiments. 

\textbf{Effectiveness of each component in Ada-CM.} To evaluate the importance of the proposed adaptive confidence margin, we carry out the ablation study to investigate the $\mathcal{L}^{u}$ with samples with high confidence scores and $\mathcal{L}^{c}$ with samples with low confidence scores on RAF-DB with 100 labels and SFEW with 400 labels. In addition, the relation in \Cref{sec:discuss} can also be verified.

\begin{table*}\footnotesize
  \centering
  \caption{Performance comparison with the state-of-the-art SSL methods on RAF-DB, SFEW and AffectNet using ResNet-18 (in \%, mean $\pm$ standard deviation). Fully supervised denotes that all labeled training data is used to train the DFER model. This also applies to the following tables. The fully-supervised baseline results are obtained by DLP-CNN \cite{Li_2019_TIP} on RAF-DB and SFEW, RAN \cite{wang2020region} on AffectNet.}
  \vspace{-3.6mm}
  \begin{tabular}{c|cccc|cc|cc}
    \toprule[1pt]
    \multirow{2}{*}{Method} & \multicolumn{4}{c|}{RAF-DB} &\multicolumn{2}{c|}{SFEW}&\multicolumn{2}{c}{AffectNet}\\\cline{2-5} \cline{6-7} \cline{8-9} 
    ~ &100 labels&400 labels&2000 labels&4000 labels&100 labels&400 labels&2000 labels&10000 labels\\
    \hline
    Baseline & 52.43{\tiny$\pm$2.24} & 67.75{\tiny$\pm$0.95} & 78.91{\tiny$\pm$0.43} & 81.90{\tiny$\pm$0.48} & 33.76{\tiny$\pm$1.84} & 43.85{\tiny$\pm$2.83} & 47.52{\tiny$\pm$0.75} & 53.18{\tiny$\pm$0.68} \\
    Pseudo-Labeling \cite{lee2013pseudo} & 54.96{\tiny$\pm$4.24} & 69.99{\tiny$\pm$1.81} & 79.18{\tiny$\pm$0.27} & 82.88{\tiny$\pm$0.49} & 34.27{\tiny$\pm$1.67} & 45.27{\tiny$\pm$1.32} & 48.78{\tiny$\pm$0.67} & 53.82{\tiny$\pm$1.29}\\
    MixMatch \cite{NEURIPS2019_1cd138d0} & 54.57{\tiny$\pm$4.16} & 73.14{\tiny$\pm$1.40} & 79.63{\tiny$\pm$0.91} & 83.57{\tiny$\pm$0.49} & 34.13{\tiny$\pm$2.58}&44.91{\tiny$\pm$1.87} &49.63{\tiny$\pm$0.49} & 53.49{\tiny$\pm$0.47}\\
    UDA \cite{NEURIPS2020_44feb009} & 58.15{\tiny$\pm$1.54} & 72.39{\tiny$\pm$1.64} & 81.16{\tiny$\pm$0.54} & 83.56{\tiny$\pm$0.82} & 39.22{\tiny$\pm$2.30} & 48.90{\tiny$\pm$1.56} & 50.42{\tiny$\pm$0.45} & 56.49{\tiny$\pm$0.27}\\
    ReMixMatch \cite{Berthelot2020ReMixMatch} & 58.83{\tiny$\pm$2.34} & 73.34{\tiny$\pm$1.82} & 79.66{\tiny$\pm$0.66} &  83.51{\tiny$\pm$0.18}  & 35.69{\tiny$\pm$2.73}&48.39{\tiny$\pm$0.71}& 50.38{\tiny$\pm$0.63}& 55.81{\tiny$\pm$0.34}\\
    FixMatch \cite{NEURIPS2020_06964dce} & 60.67{\tiny$\pm$2.25} & 73.36{\tiny$\pm$1.59} & 81.27{\tiny$\pm$0.27} & 83.31{\tiny$\pm$0.33} & 38.90{\tiny$\pm$1.90} & 50.37{\tiny$\pm$0.45} & 50.79{\tiny$\pm$0.37} & 56.50{\tiny$\pm$0.43}\\
    \textbf{Ada-CM} &\textbf{62.36}{\tiny$\pm$1.10}&\textbf{74.44}{\tiny$\pm$1.53}&\textbf{82.05}{\tiny$\pm$0.22}&\textbf{84.42}{\tiny$\pm$0.49}&\textbf{41.88}{\tiny$\pm$2.12}&\textbf{52.43}{\tiny$\pm$0.67}&\textbf{51.22}{\tiny$\pm$0.29}&\textbf{57.42}{\tiny$\pm$0.43}\\
    \hline
    Fully Supervised &\multicolumn{4}{c|}{84.13} & \multicolumn{2}{c|}{51.05}& \multicolumn{2}{c}{52.97}\\
    \bottomrule[1pt]
    \end{tabular}
    \vspace{-0.3cm}
  \label{tab:ss}
\end{table*}

As shown in \Cref{tab:ab1}, several observations can be summarized as follows. Firstly, compared with the baseline, other methods (rows 2 to 8) leverage unlabeled samples and significantly improve the baseline performance on two evaluation schemes. In all cases, our final Ada-CM (row 8) achieves the best performance improvement. Moreover, different fixed thresholds affect the quality of pseudo labels, which is consistent with the effect in FixMatch \cite{NEURIPS2020_06964dce}.

\begin{figure}[t]
  \centering
  \setlength{\abovecaptionskip}{0.1cm}
   \includegraphics[width=1\linewidth]{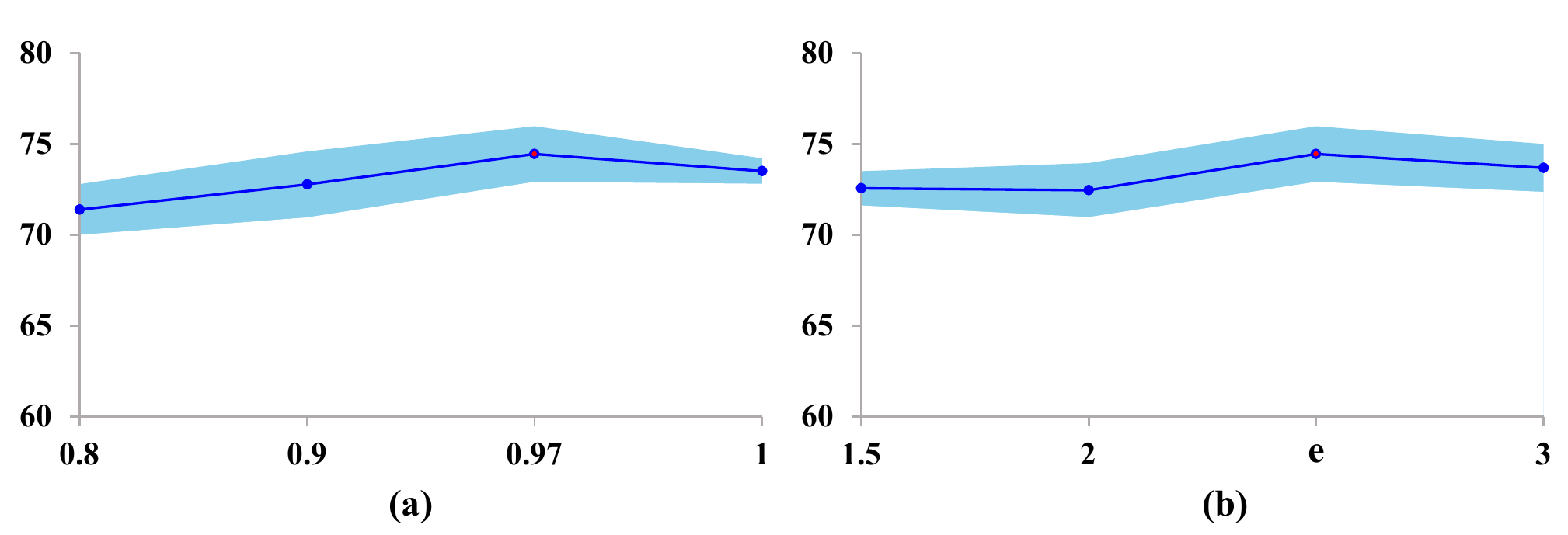}
   \caption{Plots of ablation study on Ada-CM. (a) Varying the control parameter $B$. (b) Measuring the effect of $\gamma$. The performance with default setting is marked in the red.}
   \vspace{-0.3cm}
   \label{fig:3}
\end{figure}

Secondly, the effect of the contrastive objective (row 7) exceeds the baseline but is not satisfactory. This might be explained by the reason that the contrastive objective focuses on the feature-level similarity between different views of the same data, while having limited ability to distinguish inter-class samples. However, the operation can ensure that all unlabeled samples are leveraged to update models and achieve synergy with $\mathcal{L}^{u}$ to improve performance.

In addition, for the effect of thresholds, we compare three fixed thresholds, FlexMatch (row 5) and our adaptive confidence margin (row 6). From the results, our adaptive confidence margin is shown to achieve larger performance improvement. These results validate two contributions of our method: 1) Compared with the fixed threshold-based methods, our method is highly effective in pseudo-labeling unlabeled facial expressions. 2) Our Ada-CM and FlexMatch \cite{zhang2021flexmatch} both on  samples with high confidence scores achieve similar performance. However, the contribution of our method is that the Ada-CM can leverage all unlabeled samples, compared with selecting only parts of samples in FlexMatch. Indeed, combining the adaptive confidence margin and contrastive objective, our method (row 8) achieves the best results, which demonstrates that with the help of all unlabeled data, the entropy minimization and the contrastive learning can jointly guide models to extract more discriminative features.

\begin{table}\footnotesize
  \centering
  \caption{Performance comparison with the state-of-the-art SS-DFER methods on RAF-DB using WideResNet-28-2 (in \%, mean $\pm$ standard deviation).}
  \vspace{-3.6mm}
  \begin{tabular}{c|ccc}
    \toprule[1pt]
    \multirow{2}{*}{Method} & \multicolumn{3}{c}{Labeled samples} \\\cline{2-4}
    ~ &400 &1000 &4000 \\
    \hline
    Baseline &26.75&35.25&55.66\\
    MeanTeacher \cite{NIPS2017_68053af2} &28.23&36.53&60.36\\
    MixMatch \cite{NEURIPS2019_1cd138d0} &42.25&60.37&65.24\\
    MarginMix \cite{margin2020eccv} &45.75&66.47&70.68\\
    \hline
    \textbf{Ada-CM} &\textbf{59.03}{\tiny$\pm$0.73}&\textbf{68.38}{\tiny$\pm$0.44}&\textbf{75.98}{\tiny$\pm$0.41}\\
    \bottomrule[1pt]
    \end{tabular}
    \vspace{-0.3cm}
  \label{tab:ssfer}
\end{table}

\textbf{Evaluation of $B$.} Since the parameter $B$ is used to control the peak of confidence margin at each epoch, we conduct experiments to explore different $B$ in \cref{eq:5}. \Cref{fig:3} (a) reflects the model performance with different $B$. We find that the default setting $B=0.97$ achieves the best result. When $B$ is too small, it is difficult for our method to ensure the quality of pseudo labels. The reason is that the amount of data with wrong pseudo labels increases. 

\textbf{Influence of different $\gamma$.} $\gamma$ provides the ability to gradually modify the current confidence margin. \Cref{fig:3} (b) shows the effects of different $\gamma \in \{1.5,2,e,3\}$. We can obtain that our method is not sensitive to $\gamma$ in a certain range but obtains the top performance when $\gamma$ is set to $e$.

\subsection{Comparison with State-of-the-Art Methods}
\label{sec:sota}

\begin{figure*}[t]
  \centering
  \setlength{\abovecaptionskip}{0.1cm}
  \includegraphics[width=1\linewidth]{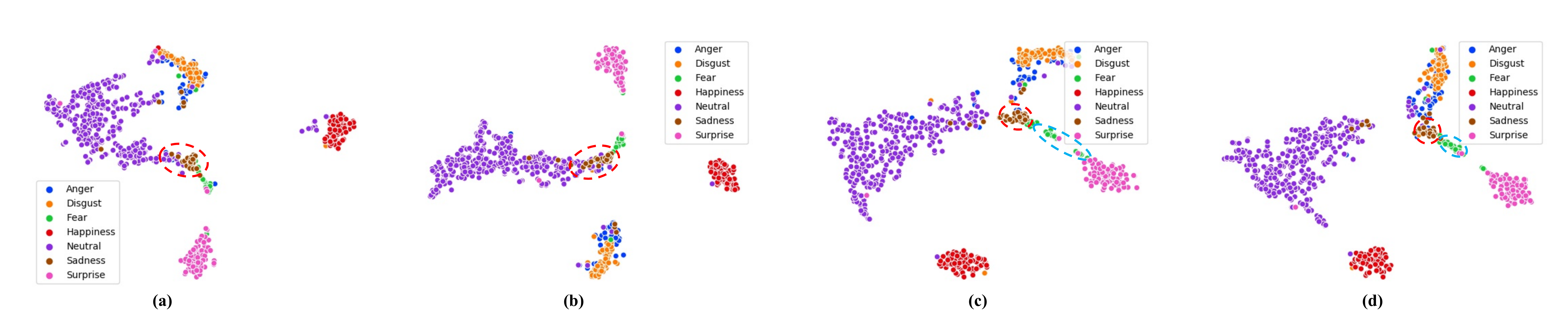}
  \caption{2D t-SNE visualization \cite{maaten2008visualizing} of facial expression features obtained by different methods, including (a) Baseline, (b) FixMatch, (c) our Ada-CM (w/o the contrastive objective) and (d) the whole Ada-CM. All models are trained on RAF-DB with 4,000 labels. The features are extracted from the CK+ dataset.}
  \vspace{-0.3cm}
  \label{fig:4}
\end{figure*}

To verify the effectiveness of our Ada-CM, we provide experimental results on RAF-DB, SFEW and AffectNet datasets to compare with state-of-the-art methods in two aspects, including comparison with the SS-DFER method \cite{margin2020eccv} on RAF-DB and comparison with SSL methods. \Cref{tab:ss} compares our method with SSL methods using ResNet-18 as the backbone network. From this table, it clearly shows that our proposed Ada-CM achieves the best performance and surpasses the state-of-the-art FixMatch \cite{NEURIPS2020_06964dce} with a large margin. This indicates that our method can better leverage unlabeled data to further improve SSL performance. Compared with the fully-supervised results \cite{Li_2019_TIP,wang2020region}, our method can still beat the baselines with large gains, \ie, 0.29\% on RAF-DB, 1.38\% on SFEW and 4.45\% on AffectNet for the case of 1/3, 1/2 and 1/28 labeled data ratio, respectively. These results verify the effectiveness of our method and the ability to deal with the real-world limited labeled case.

Besides, the proposed Ada-CM can always outperform MarginMix \cite{margin2020eccv} in each case. To the best of our knowledge, MarginMix could be the first attempt to solve the SS-DFER problem based on MixMatch \cite{NEURIPS2019_1cd138d0}. As shown in \Cref{tab:ssfer}, our Ada-CM significantly surpasses it by 13.28\%, 1.91\% and 5.3\% with 400, 1,000 and 4,000 labeled samples, respectively. The remarkable results demonstrate the effectiveness of our proposed Ada-CM in dealing with SS-DFER. More results could be found in the supplementary material.

\begin{table}\footnotesize
  \centering
  \caption{Cross-dataset evaluation on in-the-lab CK+ using WideResNet (WRN)-28-2 \cite{Zagoruyko2016WRN} and ResNet-18 \cite{he2016deep} (in \%, mean $\pm$ standard deviation). All models are trained on RAF-DB and tested on CK+ dataset.}
  \vspace{-3.6mm}
  \begin{tabular}{c|cc|c}
    \toprule[1pt]
    \multirow{2}{*}{Method} & \multicolumn{2}{c|}{Labeled samples}&\multirow{2}{*}{Backbone} \\\cline{2-3}
    ~ &100 &4000 \\
    \hline
    Baseline &44.29{\tiny$\pm$3.04} &70.97{\tiny$\pm$2.21} &\multirow{4}{*}{WRN-28-2}\\
    MixMatch \cite{NEURIPS2019_1cd138d0} & 50.42{\tiny$\pm$8.36}& 71.76{\tiny$\pm$1.48}&\\
    FixMatch \cite{NEURIPS2020_06964dce} &52.52{\tiny$\pm$9.69} & 76.98{\tiny$\pm$2.15}&\\
    \textbf{Ada-CM} &\textbf{56.13}{\tiny$\pm$6.85}&\textbf{79.34}{\tiny$\pm$1.14}&\\
    \hline
    Baseline &59.02{\tiny$\pm$3.63} & 80.63{\tiny$\pm$0.62}&\multirow{4}{*}{ResNet-18}\\
    MixMatch \cite{NEURIPS2019_1cd138d0} &59.94{\tiny$\pm$5.46} & 83.87{\tiny$\pm$1.02}&\\
    FixMatch \cite{NEURIPS2020_06964dce} &73.62{\tiny$\pm$1.78} &84.18{\tiny$\pm$0.99} &\\
    \textbf{Ada-CM} &\textbf{76.92}{\tiny$\pm$3.57}&\textbf{85.32}{\tiny$\pm$0.98}&\\
    \hline
    \multirow{2}{*}{Fully Supervised} &\multicolumn{3}{c}{81.07 \cite{li2019occlusion}} \\&\multicolumn{3}{c}{81.72 \cite{spwfa2020}} \\
    \bottomrule[1pt]
    \end{tabular}
    \vspace{-0.3cm}
  \label{tab:cross}
\end{table}

\subsection{SSL for Cross-Dataset Evaluation}
\label{sec:cross}

To further verify the generalization ability of our method, we conduct a cross-dataset evaluation scheme (RAF-DB to CK+ dataset), which is widely used in the cross-dataset DFER. \Cref{tab:cross} shows the comparison with state-of-the-art methods using WideResNet-28-2 and ResNet-18 as the backbone. Obviously, our method achieves better performance than existing methods in all cases. Compared with the fully-supervised results \cite{spwfa2020}, the Ada-CM with 4,000 labeled samples using ResNet-18 obtains larger gains by 3.6\%. It suggests that our method focuses on a large amount of unlabeled data without the influence of original labels, which is conducive to generalization. Furthermore, our method can achieve superior performance with 1/3 labeled data and fewer model parameters. To be specific, the backbone used in \cite{spwfa2020} is ResNet-50 with channel level attention, while we use the more lightweight ResNet-18. 

\subsection{Visualization}
\label{sec:visual}

To further evaluate the effectiveness of the important adaptive confidence margin in our method, we use t-SNE \cite{maaten2008visualizing} to visualize the facial expression feature distribution extracted by the baseline, FixMatch, our proposed Ada-CM (w/o the contrastive objective) and the whole Ada-CM on the 2D space, respectively. 

As shown in \Cref{fig:4}, we can observe that the facial expression features obtained by the baseline and FixMatch are not enough discriminative for some categories, \eg, the sadness in the red dotted line. In contrast, our Ada-CM (w/o the contrastive objective) can achieve a clear boundary between the sadness and other categories. Especially, after combining the contrastive objective, the intra-class similarity and inter-class differences are more distinct.

\section{Conclusion}
\label{sec:conclusion}

In this paper, we propose a novel Adaptive Confidence Margin (Ada-CM) for semi-supervised deep facial expression recognition, which adaptively leverages all unlabeled samples (\ie, samples in subset I with high confidence scores and samples in subset II with low confidence scores) to train models. The proposed Ada-CM dramatically improves the performance from two aspects. On one hand, unlabeled samples whose confidence scores exceed the learned confidence margin are directly pseudo-labeled to match the predictions of strongly-augmented versions. On the other hand, the contrastive objective is applied to learn facial expression features among samples in subset II. Experiments on four popular datasets show the superiority of our method to perform the SS-DFER task.

\noindent
\textbf{Acknowledgments:} This work was supported in part by the National Key Research and Development Program of China under Grant 2018AAA0103202, in part by the National Natural Science Foundation of China under Grant 61922066, 61876142, 62036007 and 61976166, in part by the Technology Innovation Leading Program of Shaanxi under Grant 2022QFY01-15, in part by Open Research Projects of Zhejiang Lab under Grant 2021KG0AB01, in part by the Fundamental Research Funds for the Central Universities, and in part by the Innovation Fund of Xidian University.

{\small
\bibliographystyle{ieee_fullname}
\bibliography{egbib}

\begin{thebibliography}{10}\itemsep=-1pt

\bibitem{BarsoumICMI2016}
Emad Barsoum, Cha Zhang, Cristian~Canton Ferrer, and Zhengyou Zhang.
\newblock Training deep networks for facial expression recognition with
  crowd-sourced label distribution.
\newblock In {\em ACM ICMI}, pages 279--283, 2016.

\bibitem{Berthelot2020ReMixMatch}
David Berthelot, Nicholas Carlini, Ekin~D. Cubuk, Alex Kurakin, Kihyuk Sohn,
  Han Zhang, and Colin Raffel.
\newblock Remixmatch: Semi-supervised learning with distribution matching and
  augmentation anchoring.
\newblock In {\em ICLR}, 2020.

\bibitem{NEURIPS2019_1cd138d0}
David Berthelot, Nicholas Carlini, Ian Goodfellow, Nicolas Papernot, Avital
  Oliver, and Colin~A Raffel.
\newblock Mixmatch: A holistic approach to semi-supervised learning.
\newblock In {\em NeurIPS}, volume~32, pages 5050--5060, 2019.

\bibitem{chen2020simple}
Ting Chen, Simon Kornblith, Mohammad Norouzi, and Geoffrey Hinton.
\newblock A simple framework for contrastive learning of visual
  representations.
\newblock In {\em ICML}, pages 1597--1607, 2020.

\bibitem{cubuk2020randaugment}
Ekin~D Cubuk, Barret Zoph, Jonathon Shlens, and Quoc~V Le.
\newblock Randaugment: Practical automated data augmentation with a reduced
  search space.
\newblock In {\em CVPR Workshops}, pages 702--703, 2020.

\bibitem{dhall2011static}
Abhinav Dhall, Roland Goecke, Simon Lucey, and Tom Gedeon.
\newblock Static facial expression analysis in tough conditions: Data,
  evaluation protocol and benchmark.
\newblock In {\em ICCV Workshops}, pages 2106--2112, 2011.

\bibitem{margin2020eccv}
Corneliu Florea, Mihai Badea, Laura Florea, Andrei Racoviteanu, and Constantin
  Vertan.
\newblock Margin-mix: Semi-supervised learning for face expression recognition.
\newblock In {\em ECCV}, pages 1--17, 2020.

\bibitem{grandvalet2004semi}
Yves Grandvalet and Yoshua Bengio.
\newblock Semi-supervised learning by entropy minimization.
\newblock In {\em NeurIPS}, pages 529--536, 2005.

\bibitem{he2016deep}
Kaiming He, Xiangyu Zhang, Shaoqing Ren, and Jian Sun.
\newblock Deep residual learning for image recognition.
\newblock In {\em CVPR}, pages 770--778, 2016.

\bibitem{hinton2015distilling}
Geoffrey Hinton, Oriol Vinyals, and Jeff Dean.
\newblock Distilling the knowledge in a neural network.
\newblock {\em arXiv preprint arXiv:1503.02531}, 2015.

\bibitem{hu2008multi}
Yuxiao Hu, Zhihong Zeng, Lijun Yin, Xiaozhou Wei, Xi Zhou, and Thomas~S Huang.
\newblock Multi-view facial expression recognition.
\newblock In {\em FG}, pages 1--6, 2008.

\bibitem{iscen2019label}
Ahmet Iscen, Giorgos Tolias, Yannis Avrithis, and Ondrej Chum.
\newblock Label propagation for deep semi-supervised learning.
\newblock In {\em CVPR}, pages 5070--5079, 2019.

\bibitem{kingma2014adam}
Diederik~P Kingma and Jimmy Ba.
\newblock Adam: A method for stochastic optimization.
\newblock {\em arXiv preprint arXiv:1412.6980}, 2014.

\bibitem{lee2013pseudo}
Dong-Hyun Lee.
\newblock Pseudo-label: The simple and efficient semi-supervised learning
  method for deep neural networks.
\newblock In {\em ICML Workshops}, 2013.

\bibitem{li2021tip}
Hangyu Li, Nannan Wang, Xinpeng Ding, Xi Yang, and Xinbo Gao.
\newblock Adaptively learning facial expression representation via c-f labels
  and distillation.
\newblock {\em IEEE Transactions on Image Processing}, 30:2016--2028, 2021.

\bibitem{Li_2019_TIP}
Shan Li and Weihong Deng.
\newblock Reliable crowdsourcing and deep locality-preserving learning for
  unconstrained facial expression recognition.
\newblock {\em IEEE Transactions on Image Processing}, 28(1):356--370, 2019.

\bibitem{Li_Peng_2021}
Yunfan Li, Peng Hu, Zitao Liu, Dezhong Peng, Joey~Tianyi Zhou, and Xi Peng.
\newblock Contrastive clustering.
\newblock In {\em AAAI}, volume~35, pages 8547--8555, 2021.

\bibitem{spwfa2020}
Yingjian Li, Guangming Lu, Jinxing Li, Zheng Zhang, and David Zhang.
\newblock Facial expression recognition in the wild using multi-level features
  and attention mechanisms.
\newblock {\em IEEE Transactions on Affective Computing}, 2020.

\bibitem{li2019occlusion}
Yong Li, Jiabei Zeng, Shiguang Shan, and Xilin Chen.
\newblock Occlusion aware facial expression recognition using cnn with
  attention mechanism.
\newblock {\em IEEE Transactions on Image Processing}, 28(5):2439--2450, 2019.

\bibitem{lucey2010extended}
Patrick Lucey, Jeffrey~F. Cohn, Takeo Kanade, Jason Saragih, Zara Ambadar, and
  Iain Matthews.
\newblock The extended cohn-kanade dataset (ck+): A complete dataset for action
  unit and emotion-specified expression.
\newblock In {\em CVPR Workshops}, pages 94--101, 2010.

\bibitem{luo2013facial}
Yuan Luo, Cai-ming Wu, and Yi Zhang.
\newblock Facial expression recognition based on fusion feature of pca and lbp
  with svm.
\newblock {\em Optik-International Journal for Light and Electron Optics},
  124(17):2767--2770, 2013.

\bibitem{mollahosseini2017affectnet}
Ali Mollahosseini, Behzad Hasani, and Mohammad~H Mahoor.
\newblock Affectnet: A database for facial expression, valence, and arousal
  computing in the wild.
\newblock {\em IEEE Transactions on Affective Computing}, 10(1):18--31, 2017.

\bibitem{pietikainen2011computer}
Matti Pietik{\"a}inen, Abdenour Hadid, Guoying Zhao, and Timo Ahonen.
\newblock {\em Computer vision using local binary patterns}, volume~40.
\newblock Springer Science \& Business Media, 2011.

\bibitem{NIPS2015_378a063b}
Antti Rasmus, Mathias Berglund, Mikko Honkala, Harri Valpola, and Tapani Raiko.
\newblock Semi-supervised learning with ladder networks.
\newblock In {\em NeurIPS}, volume~28, page 3546–3554, 2015.

\bibitem{ssl_conf_05}
Chuck Rosenberg, Martial Hebert, and Henry Schneiderman.
\newblock Semi-supervised self-training of object detection models.
\newblock In {\em WACV/MOTION}, volume~1, pages 29--36, 2005.

\bibitem{sajjadi2016regularization}
Mehdi Sajjadi, Mehran Javanmardi, and Tolga Tasdizen.
\newblock Regularization with stochastic transformations and perturbations for
  deep semi-supervised learning.
\newblock In {\em NeurIPS}, volume~29, pages 1163--1171, 2016.

\bibitem{she2021dive}
Jiahui She, Yibo Hu, Hailin Shi, Jun Wang, Qiu Shen, and Tao Mei.
\newblock Dive into ambiguity: Latent distribution mining and pairwise
  uncertainty estimation for facial expression recognition.
\newblock In {\em CVPR}, pages 6248--6257, 2021.

\bibitem{NEURIPS2020_06964dce}
Kihyuk Sohn, David Berthelot, Nicholas Carlini, Zizhao Zhang, Han Zhang,
  Colin~A Raffel, Ekin~Dogus Cubuk, Alexey Kurakin, and Chun-Liang Li.
\newblock Fixmatch: Simplifying semi-supervised learning with consistency and
  confidence.
\newblock In {\em NeurIPS}, volume~33, pages 596--608, 2020.

\bibitem{NIPS2017_68053af2}
Antti Tarvainen and Harri Valpola.
\newblock Mean teachers are better role models: Weight-averaged consistency
  targets improve semi-supervised deep learning results.
\newblock In {\em NeurIPS}, volume~30, pages 1195--1204, 2017.

\bibitem{maaten2008visualizing}
Laurens Van~der Maaten and Geoffrey Hinton.
\newblock Visualizing data using t-sne.
\newblock {\em Journal of machine learning research}, 9(11):2579--2605, 2008.

\bibitem{wang2020suppressing}
Kai Wang, Xiaojiang Peng, Jianfei Yang, Shijian Lu, and Yu Qiao.
\newblock Suppressing uncertainties for large-scale facial expression
  recognition.
\newblock In {\em CVPR}, pages 6897--6906, 2020.

\bibitem{wang2020region}
Kai Wang, Xiaojiang Peng, Jianfei Yang, Debin Meng, and Yu Qiao.
\newblock Region attention networks for pose and occlusion robust facial
  expression recognition.
\newblock {\em IEEE Transactions on Image Processing}, 29:4057--4069, 2020.

\bibitem{wang2021data}
Zhenyu Wang, Yali Li, Ye Guo, Lu Fang, and Shengjin Wang.
\newblock Data-uncertainty guided multi-phase learning for semi-supervised
  object detection.
\newblock In {\em CVPR}, pages 4568--4577, 2021.

\bibitem{NEURIPS2020_44feb009}
Qizhe Xie, Zihang Dai, Eduard Hovy, Thang Luong, and Quoc Le.
\newblock Unsupervised data augmentation for consistency training.
\newblock In {\em NeurIPS}, volume~33, pages 6256--6268, 2020.

\bibitem{xu2021dash}
Yi Xu, Lei Shang, Jinxing Ye, Qi Qian, Yu-Feng Li, Baigui Sun, Hao Li, and Rong
  Jin.
\newblock Dash: Semi-supervised learning with dynamic thresholding.
\newblock In {\em ICML}, pages 11525--11536, 2021.

\bibitem{Xue_2021_ICCV}
Fanglei Xue, Qiangchang Wang, and Guodong Guo.
\newblock Transfer: Learning relation-aware facial expression representations
  with transformers.
\newblock In {\em ICCV}, pages 3601--3610, 2021.

\bibitem{yao2021jo}
Yazhou Yao, Zeren Sun, Chuanyi Zhang, Fumin Shen, Qi Wu, Jian Zhang, and
  Zhenmin Tang.
\newblock Jo-src: A contrastive approach for combating noisy labels.
\newblock In {\em CVPR}, pages 5192--5201, 2021.

\bibitem{Zagoruyko2016WRN}
Sergey Zagoruyko and Nikos Komodakis.
\newblock Wide residual networks.
\newblock In {\em BMVC}, pages 87.1--87.12, 2016.

\bibitem{zeng2018facial}
Jiabei Zeng, Shiguang Shan, and Xilin Chen.
\newblock Facial expression recognition with inconsistently annotated datasets.
\newblock In {\em ECCV}, pages 222--237, 2018.

\bibitem{zhang2021flexmatch}
Bowen Zhang, Yidong Wang, Wenxin Hou, Hao Wu, Jindong Wang, Manabu Okumura, and
  Takahiro Shinozaki.
\newblock Flexmatch: Boosting semi-supervised learning with curriculum pseudo
  labeling.
\newblock In {\em NeurIPS}, volume~34, 2021.

\bibitem{zhang2016joint}
Kaipeng Zhang, Zhanpeng Zhang, Zhifeng Li, and Yu Qiao.
\newblock Joint face detection and alignment using multitask cascaded
  convolutional networks.
\newblock {\em IEEE Signal Processing Letters}, 23(10):1499--1503, 2016.

\bibitem{zhao2011facial}
Guoying Zhao, Xiaohua Huang, Matti Taini, Stan~Z Li, and Matti Pietik{\"a}Inen.
\newblock Facial expression recognition from near-infrared videos.
\newblock {\em Image and Vision Computing}, 29(9):607--619, 2011.

\end{thebibliography}
}

\end{document}